\documentclass{article}
\usepackage{spconf,amsmath,graphicx,hyperref,booktabs}
\usepackage{marginnote}
\usepackage{setspace} 
\usepackage{comment}

\title{A Computational Approach to Analyzing Disrupted Language in Schizophrenia: Integrating Surprisal and Coherence Measures}
\name{Gowtham Premananth, Carol Espy-Wilson\thanks{This work was supported by the National Science Foundation grant numbered 2124270.}}
\address{Institute for System Research, Department of Electrical and Computer Engineering,\\  
University of Maryland, College Park, USA\\
}
\begin{document}
%
\maketitle
\begin{abstract}
Language disruptions are one of the well-known effects of schizophrenia symptoms. They are often manifested as disorganized speech and impaired discourse coherence. These abnormalities in spontaneous language production reflect underlying cognitive disturbances and have the potential to serve as objective markers for symptom severity and diagnosis of schizophrenia. This study focuses on how these language disruptions can be characterized in terms of  two computational linguistic measures: surprisal and semantic coherence. By computing surprisal and semantic coherence of language  using computational models, this study investigates how they differ between subjects with schizophrenia and healthy controls. Furthermore, this study provides further insight into how language disruptions in terms of these linguistic measures change with varying degrees of schizophrenia symptom severity.
\end{abstract}
\vspace{-0.5cm}
\section{Introduction}
\vspace{-0.3cm}
Schizophrenia is a complex mental health disorder that is found in around one percent of the world population \cite{institute2021global}. Schizophrenia subjects exhibit a wide range of symptoms, like hallucinations, delusions, diminished emotional expression, and disorganized thinking. Most of the symptoms that are usually exhibited by schizophrenia subjects have been known to affect how speech and language is produced \cite{kuperberg2010language}. Therefore different natural language processing techniques and language modeling techniques have been used to detect and estimate the disturbance caused by language production among schizophrenia subjects \cite{jeong2023exploring}. In addition, these changes are being incorporated into multimodal systems to better distinguish schizophrenia subjects from healthy controls as well \cite{embc,premananth24_interspeech}. 

Semantic coherence in speech is one of the most commonly investigated features of speech when it comes to schizophrenia subjects \cite{just2023validation}. Semantic coherence measures how logically and meaningfully connected is the content of the produced speech. When it comes to the speech of schizophrenia subjects, they tend to show lower semantic coherence when compared to healthy controls \cite{elvevaag2007quantifying}.

Surprisal theory \cite{hale-2001-probabilistic} investigates the processing difficulty of a word or a phrase in a sentence based on their predictability with respect to the preceding context of the sentence. This quantifies how much the specific word or phrase is unexpected or informative in the context of the sentence. Even though quantifying surprisal using language models has been one of the most commonly investigated topics in computational psycholinguistics \cite{levy2008expectation,wilcox2023testing}, it hasn't been properly investigated when it comes to the language disruptions caused by schizophrenia. Also the relationship between semantic coherence and surprisal has not been investigated in the context of language disruptions caused by schizophrenia.

This study aims to investigate the language disruptions in speech associated with schizophrenia by addressing the following research questions based on spontaneous speech. We investigate whether individuals with schizophrenia exhibit higher surprisal and lower semantic coherence when compared to healthy controls. In addition, we explore the relationship between surprisal and semantic coherence in the speech of healthy individuals and whether this relationship is altered or affected in any way in subjects with schizophrenia. Finally, we assess how the severity of clinical symptoms exhibited by subjects affect the semantic coherence and surprisal in the speech of the subjects.
\vspace{-0.3cm}
\section{Dataset and Data Pre-processing}
\vspace{-0.2cm}
The dataset used in the study was collected as a part of a mental health study conducted at the school of medicine at University of Maryland in collaboration with University of Maryland College Park \cite{KELLY2020113496}. This study included subjects with a diagnosis of schizophrenia, major depressive disorder, and healthy controls. The dataset was collected in an in-person setup where the subjects met with a clinician in the school of medicine's clinic. They were given multiple clinician-assessed mental health symptomatology questionnaires like the Brief Psychiatric Rating Scale (BPRS) \cite{overall1962brief} and the Hamilton Depression rating scale (HAM-D) \cite{hamilton1960rating}. These questionnaires were used to assess the severity of different symptoms in addition to the disorder diagnosis provided by the clinicians. 

The BPRS questionnaire used in this study is an 18 item symptomatology questionnaire that rates 18 different symptoms associated with schizophrenia. Each symptom was rated on a scale of 1-7 based on the severity of the specific symptom. The BPRS scores assessed by clinicians were used as subjective measures of schizophrenia symptom severity. These scores served as the basis for experiments investigating how language disruption metrics in speech vary with symptom severity.

After the assessment the subjects took part in an interview session with an interviewer which was recorded. These interview sessions were conducted in an unstructured manner where the interview questions were not predetermined, but decided based on the answers provided by the subjects. This framework allowed the dataset to be completely made up of spontaneous conversational speech of the participants. The interview recordings were manually transcribed by a third-party transcription service.

The participants enrolled in this study attended multiple sessions up to a total of 4 sessions per each participant with each session after a week from the previous session. The overall dataset included a total of 196 sessions belonging to 57 unique participants out of which 23 had a diagnosis of schizophrenia, 16 had a diagnosis of major depressive disorder and 18 were healthy controls. For all the experiments conducted in this work, we used a total of 140 sessions belonging to 39 unique subjects with a diagnosis of either Schizophrenia (SZ) or Healthy Controls (HC). The subjects selected for this study had clinically-assessed overall BPRS scores between the range of 18-67. The subjects with major depressive disorder were excluded from the experiments in this study. 

The transcripts of the interview sessions in the dataset were used to extract the speech of the subjects in the dataset. And the extracted speech of the subjects were used as inputs for all the language models used in the experiments to calculate semantic coherence and surprisal scores. Each answer provided by the subject responding to a question by the interviewer was considered as an utterance. And all the utterances were treated as separate inputs to the models. The utterances in the dataset varied in length. The shortest utterances were one sentence utterances while the longest utterances in the dataset were nine sentence utterances. The NLTK (Natural Language Tool Kit) library's "sent\_tokenize" function was used to tokenize the sentences in the utterances before being used as the inputs to calculate surprisal and semantic coherence scores. 
\vspace{-0.5cm}
\section{Experiments}
\vspace{-0.3cm}
The experiments conducted as a part of this study mainly focused on the surprisal and semantic coherence of the speech produced. Therefore, initially semantic coherence and surprisal scores were computed for all the utterances used in this study. After computing these scores, analysis was conducted both separately and together for healthy controls and schizophrenia subjects in the dataset. 
\vspace{-0.3cm}
\subsection{Computing Surprisal}

Surprisal is defined as the negative log probability of a word conditioned on the probability of the preceding context as shown in Eq.1.
\vspace{-0.2cm}
\begin{equation}
    Surprisal(w_i) = -log P(w_i|w_1,w_2,....,w_{i-1})
\end{equation}

Initial works on modeling surprisal were based on probabilistic models \cite{hale-2001-probabilistic, levy2008expectation}. Currently, with the recent development of transformer-based language models and large language models, probabilistic models are replaced by them used for computing surprisal scores \cite{michaelov2022more}. Therefore, for this study the surprisal of each word in the speech of the subjects was calculated using GPT-2 model's next word probability. In order to compute the surprisal, the text is tokenized and fed into the GPT-2 model. The model outputs the probability distributions of the word based on the overall vocabulary the model was trained on. The surprisal score is then derived from the probability predicted by the model given the probability of all the preceding words predicted by the model. Based on the word-level surprisal scores, sentence-level and utterance-level scores were derived by summing all the individual word level surprisal scores.

\begin{table*}[th!]
\begin{center} 
\caption{Diagnosis-based average scores} 
\label{score_table} 
\vskip 0.15in
\begin{tabular}{l c c} 
\hline
Average Scores   &  Healthy controls & Schizophrenia subjects \\
\hline
\hline
Surprisal (Sentence-wise) & 50.45 & 50.98  \\
Semantic Coherence (LDA based) & 0.39  & 0.37\\
Semantic Coherence (BERT based)  &0.28 & 0.27  \\
\hline
\end{tabular} 
\end{center} 
\end{table*}

\vspace{-0.3cm}
\subsection {Computing Semantic Coherence}

Latent Dirichlet Allocation (LDA) \cite{blei2003latent} and BERT \cite {devlin2019bert} embeddings-based approaches were used to derive semantic coherence scores across utterances. This dual-method approach was selected to guarantee robustness in capturing various aspects of coherence and to capitalize on the complementary strengths of each model.

The consistency of topic distributions across text segments is the basis for quantifying coherence using LDA, which is a probabilistic topic modeling framework. Because it models semantic similarity in terms of shared topical content, this approach is especially well-suited for analyzing topic-level continuity. Nevertheless, LDA ignores word order and syntactic structure and treats text as a collection of words, which may reduce its sensitivity to some minute pragmatic errors.

BERT embeddings-based coherence measures, on the other hand, use contextualized sentence embeddings that are obtained from transformer architectures. By simulating the contextual and sequential dependencies both within and between sentences, these embeddings are able to capture deep semantic relationships. The BERT embeddings-based method offers a more sophisticated indicator of local semantic flow by calculating the cosine similarities between neighboring sentence embeddings. This is especially helpful in clinical narratives or dialogue, where meaning is frequently dispersed throughout sentences and greatly influenced by discourse context. 

When it comes to calculating the semantic coherence of the utterances using BERT embeddings, the number of sentences present in the utterance was needed to be taken into consideration. To check the similarity in topics between sentences we need to have multiple sentences in the utterance. To maintain this requirement we included only utterances with at least 3 sentences as inputs to calculate semantic coherence.
\vspace{-0.4cm}
\section{Results and Discussion}
\vspace{-0.3cm}

\begin{figure}[h!]
\includegraphics[width=\linewidth,height=7.3cm]{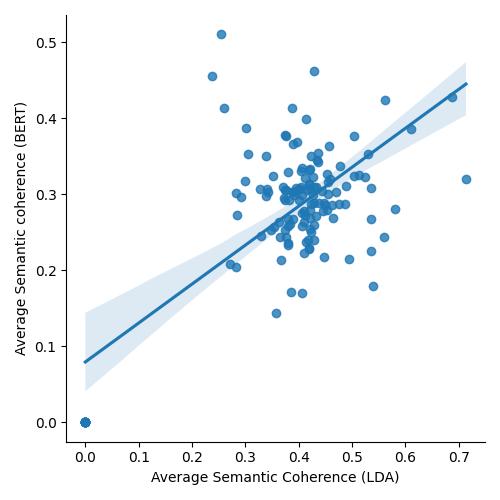}
\caption{ BERT-based semantic coherence and LDA-based semantic coherence}
\label{fig:lsc_bsc}
\end{figure}
\vspace{-0.2cm}

Table.\ref{score_table} shows the average semantic coherence and surprisal scores for all the sessions available in the dataset based on the diagnosis provided. From these results it is evident that the schizophrenia subjects have a slightly lower semantic coherence score based on both the BERT-based and LDA-based computations when compared to healthy controls. And when it comes to average surprisal schizophrenia subjects show slightly elevated scores when compared to healthy controls. The difference between the healthy controls and schizophrenia subjects based on the surprisal and semantic coherence metrics is small. One of the reasons for this might be because the schizophrenia subjects that are included in this dataset are normally functioning individuals. So these schizophrenia subjects don't show any extremely severe schizophrenia symptoms but mostly mild to moderate schizophrenia symptoms. Wider difference between healthy controls and schizophrenia subjects could be observed if subjects with extreme symptoms were available in this dataset. However, if hospitalized subjects with extreme symptoms were considered then the results will be hugely biased because of the medication used to treat extreme schizophrenia symptoms.

Next, an analysis was performed to find out how the semantic coherence scores obtained using LDA-based approach and the BERT-based approach performed on the datasets. Even though the semantic coherence scores for the same subjects from the two approaches were not the same, they were directly proportional to each other. This trend can be seen in Figure.\ref{fig:lsc_bsc}. Based on these results for all the following analysis the semantic coherence scores from the BERT embeddings-based approach was used.

\begin{figure}[h!]
\includegraphics[width=\linewidth]{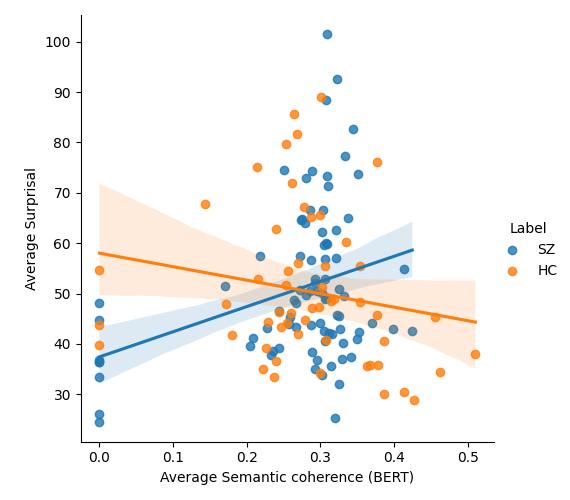}
\caption{ Surprisal vs semantic coherence}
\label{fig:surprisal_bsc}
\end{figure}
\vspace{-0.2cm}
The next set of analysis focused on observing the relation between surprisal and semantic coherence scores of all the subjects in the dataset. Figure.\ref{fig:surprisal_bsc} shows the relation between BERT-based semantic coherence and surprisal scores. As it can be seen from this figure the relation between semantic coherence and surprisal is inversely proportional for healthy controls while semantic coherence is directly proportional to surprisal scores for schizophrenia subjects.

\begin{figure}[h!]
\includegraphics[width=\linewidth,height=7.3cm]{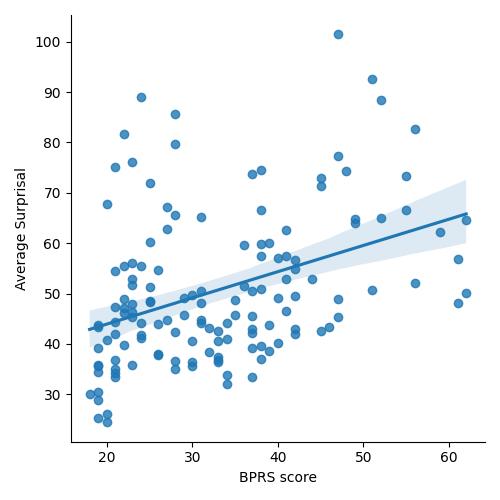}
\caption{Average surprisal in relation to symptom severity based on BPRS scores}
\label{fig:BPRS_surprisal}
\end{figure}

After analyzing the different trends shown based on the diagnosis, analysis based on how surprisal and semantic coherence varies based on the severity of the symptoms was conducted. These analysis were conducted without considering the diagnosis but by only considering the BPRS-based overall severity scores provided by the clinically assessed questionnaires. Figure.\ref{fig:BPRS_surprisal} shows how the surprisal scores vary according to the overall BPRS scores. And from the figure it is evident that the average surprisal scores increase with the increase of BPRS scores. This shows that the subjects with severe schizophrenia symptoms show higher surprisal than subjects with lower schizophrenia symptoms.
\vspace{-0.2cm}

\begin{figure}[h!]
\includegraphics[width=\linewidth,height=7.3cm]{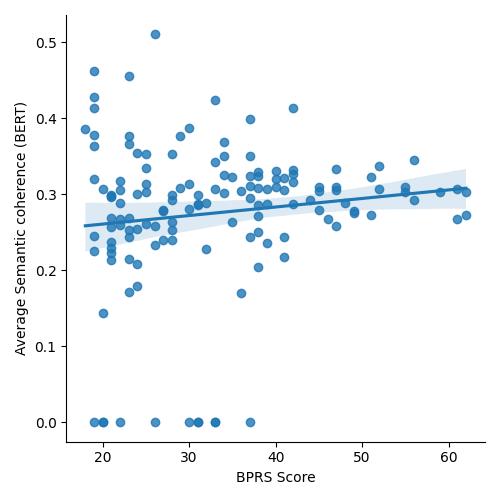}
\caption{Semantic coherence based on BERT-embeddings in relation to symptom severity based on BPRS scores}
\label{fig:BPRS_sc}
\end{figure}
\vspace{-0.2cm}

Similar analysis was conducted on the semantic coherence based on BERT-embeddings, to check how they varied with overall BPRS scores. And the results of this analysis is shown in Figure.\ref{fig:BPRS_sc}. From the results in the figure, there is not much of change in the semantic coherence score as the average trend-line in the plot only varies between 0.26 and 0.31 for the whole range of the BPRS scores from 18-67. This might be because of the fact that most of the samples in the dataset had similar semantic coherence scores and an unbalance dataset with a high number of the subjects having BPRS scores in the middle and lower ends of the spectrum. Even though there is not a clear trend visible in this figure, and semantic coherence had opposing trends for schizophrenia subjects and healthy controls, an analysis done on a more diverse dataset could result in a better conclusion regarding how semantic coherence changes with varying schizophrenia symptom severity.

\vspace{-0.5cm}
\section{Conclusion \& Future work}
\vspace{-0.3cm}
This study focused on characterizing the language disruptions caused in speech by schizophrenia in terms of changes in semantic coherence and surprisal. The findings of this study based on conversational spontaneous speech obtained in a clinical setting show that schizophrenia subjects show slightly reduced semantic coherence than healthy controls. These results were consistent irrespective of the modeling approach used to compute semantic coherence scores. In addition to this, the schizophrenia subjects showed slightly elevated average surprisal scores in comparison with healthy controls. Further analysis conducted on the correlation of semantic coherence and surprisal showed that in healthy controls the surprisal was inversely proportional to the semantic coherence while in schizophrenia subjects the opposite trend was observed. These initial results were indicative of the fact that schizophrenia symptoms were causing language disruptions and making schizophrenia subjects produce speech in a different manner than healthy individuals. This hypothesis was proven by the analysis done based on the severity of the symptoms irrespective of the diagnosis where subjects with higher schizophrenia symptoms showed higher average surprisal scores which showed irregular speech with repetition of words and words being included out of context in the utterances.

All these results obtained from this study are preliminary results based on a small dataset with only normally functioning individuals with mainly mild to moderate schizophrenia symptoms. Because of this difference between healthy controls and the schizophrenia subjects in terms of language disruption was visible but not to the level of easily distinguishable. However, these findings open up a new avenue for research where more detailed studies on more diverse and larger datasets that are descriptive of the whole range of the schizophrenia spectrum could be conducted to either support or disprove these initial findings from this study.
\vspace{-0.3cm}
\bibliographystyle{IEEEbib}
\bibliography{refs}

@article{KELLY2020113496,
title = {Blinded Clinical Ratings of Social Media Data are Correlated with In-Person Clinical Ratings in Participants Diagnosed with Either Depression, Schizophrenia, or Healthy Controls},
journal = {Psychiatry Research},
volume = {294},
pages = {113496},
year = {2020},
issn = {0165-1781},
doi = {https://doi.org/10.1016/j.psychres.2020.113496},
author = {D.L. Kelly and M. Spaderna and V. Hodzic and S. Nair and C. Kitchen and A.E. Werkheiser and M.M. Powell and F. Liu and G.Coppersmith and S. Chen and P. Resnik},
}

@misc{institute2021global,
  title={Global Health Data Exchange},
  author={{Institute of Health Metrics and Evaluation}},
  year={2021},
  publisher={Institute of Health Metrics and Evaluation Seattla, WA, USA}
}

@article{kuperberg2010language,
  title={Language in schizophrenia part 1: an introduction},
  author={Kuperberg, Gina R},
  journal={Language and linguistics compass},
  volume={4},
  number={8},
  pages={576--589},
  year={2010},
  publisher={Wiley Online Library}
}

@article{jeong2023exploring,
  title={Exploring the use of natural language processing for objective assessment of disorganized speech in schizophrenia},
  author={Jeong, Lydia and Lee, Melissa and Eyre, Ben and Balagopalan, Aparna and Rudzicz, Frank and Gabilondo, Cedric},
  journal={Psychiatric Research and Clinical Practice},
  volume={5},
  number={3},
  pages={84--92},
  year={2023}
}

@article{just2023validation,
  title={Validation of natural language processing methods capturing semantic incoherence in the speech of patients with non-affective psychosis},
  author={Just, Sandra Anna and Br{\"o}cker, Anna-Lena and Ryazanskaya, Galina and Nenchev, Ivan and Schneider, Maria and Bermpohl, Felix and Heinz, Andreas and Montag, Christiane},
  journal={Frontiers in Psychiatry},
  volume={14},
  pages={1208856},
  year={2023},
  publisher={Frontiers Media SA}
}

@article{elvevaag2007quantifying,
  title={Quantifying incoherence in speech: an automated methodology and novel application to schizophrenia},
  author={Elvev{\aa}g, Brita and Foltz, Peter W and Weinberger, Daniel R and Goldberg, Terry E},
  journal={Schizophrenia research},
  volume={93},
  number={1-3},
  pages={304--316},
  year={2007},
  publisher={Elsevier}
}

@inproceedings{hale-2001-probabilistic,
    title = "A Probabilistic {E}arley Parser as a Psycholinguistic Model",
    author = "Hale, John",
    booktitle = "Second Meeting of the North {A}merican Chapter of the Association for Computational Linguistics",
    year = "2001",
    url = "https://aclanthology.org/N01-1021/"
}

@article{levy2008expectation,
  title={Expectation-based syntactic comprehension},
  author={Levy, Roger},
  journal={Cognition},
  volume={106},
  number={3},
  pages={1126--1177},
  year={2008},
  publisher={Elsevier}
}

@article{wilcox2023testing,
  title={Testing the predictions of surprisal theory in 11 languages},
  author={Wilcox, Ethan G and Pimentel, Tiago and Meister, Clara and Cotterell, Ryan and Levy, Roger P},
  journal={Transactions of the Association for Computational Linguistics},
  volume={11},
  pages={1451--1470},
  year={2023},
  publisher={MIT Press One Broadway, 12th Floor, Cambridge, Massachusetts 02142, USA~…}
}

@article{blei2003latent,
  title={Latent dirichlet allocation},
  author={Blei, David M and Ng, Andrew Y and Jordan, Michael I},
  journal={Journal of machine Learning research},
  volume={3},
  number={Jan},
  pages={993--1022},
  year={2003}
}

@inproceedings{devlin2019bert,
  title={Bert: Pre-training of deep bidirectional transformers for language understanding},
  author={Devlin, Jacob and Chang, Ming-Wei and Lee, Kenton and Toutanova, Kristina},
  booktitle={Proceedings of the 2019 conference of the North American chapter of the association for computational linguistics: human language technologies, volume 1 (long and short papers)},
  pages={4171--4186},
  year={2019}
}

@article{overall1962brief,
  title={The brief psychiatric rating scale},
  author={Overall, John E and Gorham, Donald R},
  journal={Psychological reports},
  volume={10},
  number={3},
  pages={799--812},
  year={1962},
  publisher={SAGE Publications Sage CA: Los Angeles, CA}
}

@article{hamilton1960rating,
  title={A rating scale for depression},
  author={Hamilton, Max},
  journal={Journal of neurology, neurosurgery, and psychiatry},
  volume={23},
  number={1},
  pages={56},
  year={1960}
}

@inproceedings{michaelov2022more,
  title={The more human-like the language model, the more surprisal is the best predictor of N400 amplitude},
  author={Michaelov, James and Bergen, Ben},
  booktitle={Neurips 2022 workshop on information-theoretic principles in cognitive systems},
  year={2022}
}

@inproceedings{premananth24_interspeech,
  title     = {{A Multimodal Framework for the Assessment of the Schizophrenia Spectrum}},
  author    = {Gowtham Premananth and Yashish M. Siriwardena and Philip Resnik and Sonia Bansal and Deanna L.Kelly and Carol Espy-Wilson},
  year      = {2024},
  booktitle = {{Interspeech 2024}},
  pages     = {1470--1474},
  doi       = {10.21437/Interspeech.2024-2224},
  issn      = {2958-1796},
}

@inproceedings{embc,
  title={A multi-modal approach for identifying schizophrenia using cross-modal attention},
  author={Premananth, Gowtham and Siriwarden, Yashish M and Resnik, Philip and Espy-Wilson, Carol},
  booktitle={2024 46th Annual International Conference of the IEEE Engineering in Medicine and Biology Society (EMBC)},
  pages={1--5},
  year={2024},
  organization={IEEE}
}

\end{document}